\documentclass[10pt,twocolumn,letterpaper]{article}

\usepackage{cvpr}
\usepackage{times}
\usepackage{epsfig}
\usepackage{graphicx}
\usepackage{amsmath}
\usepackage{amssymb}
\usepackage{comment}

\usepackage{color}
\usepackage{graphicx}
\usepackage{ifthen}

\newcommand{\softmax}{\mathop{\mathrm{softmax}}}

\definecolor{gray}{rgb}{0.5,0.5,0.5}
\definecolor{green}{rgb}{0, 0.4, 0}
\definecolor{orange}{rgb}{1, 0.5, 0}
\definecolor{mahogany}{rgb}{0.75, 0.25, 0.0}
\definecolor{purple}{rgb}{0.6, 0, 0.6}
\definecolor{purple}{rgb}{0.6, 0, 0.6}
\definecolor{darkgreen}{rgb}{0, 0.4, 0}
\definecolor{frenchblue}{rgb}{0.0, 0.45, 0.73}
\newboolean{revising}
\setboolean{revising}{false}
\ifthenelse{\boolean{revising}}
{
	\newcommand{\ignore}[1]{}
	\newcommand{\jc}[1]{\textcolor{frenchblue}{#1}}
	\newcommand{\jcreplace}[2]{\textcolor{frenchblue}{#2}}
	\newcommand{\hao}[1]{\textcolor{darkgreen}{#1}}
	\newcommand{\haoreplace}[2]{\textcolor{darkgreen}{#2}}
	\newcommand{\chou}[1]{\textcolor{orange}{#1}}
	\newcommand{\choureplace}[2]{\textcolor{orange}{#2}}
	\newcommand{\sun}[1]{\textcolor{purple}{#1}}
	\newcommand{\sunrep}[2]{\textcolor{purple}{#2}}
} {
	\newcommand{\ignore}[1]{}
	\newcommand{\jc}[1]{#1}
	\newcommand{\jcreplace}[2]{#2}
	\newcommand{\hao}[1]{#1}
	\newcommand{\haoreplace}[2]{#2}
	\newcommand{\chou}[1]{#1}
	\newcommand{\choureplace}[2]{#2}
	\newcommand{\sun}[1]{#1}
	\newcommand{\sunrep}[2]{#2}
}

\newcommand{\cutabstractup}{\vspace*{-0.2in}}
\newcommand{\cutabstractdown}{\vspace*{-0.2in}}

\newcommand{\cutsectionup}{\vspace*{-0.05in}}
\newcommand{\cutsectiondown}{\vspace*{-0.09in}}

\newcommand{\cutsubsectionup}{\vspace*{-0.06in}} 
\newcommand{\cutsubsectiondown}{\vspace*{-0.08in}} 

\newcommand{\cutsubsubsectionup}{\vspace*{-0.1in}} 
\newcommand{\cutsubsubsectiondown}{\vspace*{-0.08in}} 

\newcommand{\cutcaptionup}{\vspace*{-0.05in}}
\newcommand{\cutcaptiondown}{\vspace*{-0.03in}}

\newcommand{\cuttableup}{\vspace*{-0.25in}}
\newcommand{\cuttabledown}{\vspace*{-0.08in}}

\newcommand{\cutfigureup}{\vspace*{-0.15in}}
\newcommand{\cutfiguredown}{\vspace*{-0.8in}}

\usepackage[pagebackref=true,breaklinks=true,letterpaper=true,colorlinks,bookmarks=false]{hyperref}

\usepackage[breaklinks=true,bookmarks=false]{hyperref}

\cvprfinalcopy 

\def\cvprPaperID{799} 
\def\httilde{\mbox{\tt\raisebox{-.5ex}{\symbol{799}}}}

\ifcvprfinal\fi
\begin{document}

\title{Agent-Centric Risk Assessment:\\Accident Anticipation and Risky Region Localization}

\author{Kuo-Hao Zeng$^*$$^{\dagger}$\enspace Shih-Han Chou$^*$\enspace Fu-Hsiang Chan$^*$\enspace Juan Carlos Niebles$^{\dagger}$\enspace Min Sun$^*$\\
$^{\dagger}$Stanford University\enspace$^*$National Tsing Hua University\\
\{khzeng, jniebles\}@cs.stanford.edu
\enspace\{happy810705, corgi1205\}@gmail.com\enspace sunmin@ee.nthu.edu.tw
}

\maketitle
\thispagestyle{empty}

\begin{abstract}
   For survival, a living agent (e.g., human in Fig.~\ref{fig:teaser}(a)) must have the ability to assess risk (1) by temporally anticipating accidents before they occur (Fig.~\ref{fig:teaser}(b)), and (2) by spatially localizing risky regions (Fig.~\ref{fig:teaser}(c)) in the environment to move away from threats.
In this paper, we take an agent-centric approach to study the accident anticipation and risky region localization tasks.
We propose a novel soft-attention Recurrent Neural Network (RNN) which explicitly models both spatial and appearance-wise non-linear interaction between the agent triggering the event and another agent or static-region involved.
In order to test our proposed method,
we introduce the Epic Fail (EF) dataset consisting of 3000 viral videos capturing various accidents.
In the experiments, we evaluate the risk assessment accuracy both in the temporal domain (accident anticipation) and spatial domain (risky region localization) on our EF dataset and the Street Accident (SA) dataset. Our method consistently outperforms other baselines on both datasets.
\end{abstract}

\vspace{-5mm}
\cutsectionup\section{Introduction}\label{sec.Intro}\cutsectiondown


    A very important goal for living agents in the world is survival.
In order to survive, they naturally have the ability to assess risk. For instance, humans exhibit emotional responses while taking or observing risky actions \cite{pessoa2010emotion},
in an unconscious process that appears to happen without sophisticated reasoning \cite{naber2012animal}.
Furthermore, humans have the ability to turn their attention in the risky areas of the environment more often than others \cite{okonkwo2008visual}, as risk does not distribute uniformly across the environment.
Such risk localization is very important for the agent to move away to safety.
On the other hand, humans also have the ability to assess longer-term risk by imagining future situations. In this case,
high-level reasoning techniques (imagination, simulation) can be used to assess risk in a longer term.
Such anticipation ability is also critical for the agent to react before an accident occurs.
We are inspired by these key capabilities of human intelligence and perception to study the problem of risk assessment from a computer vision perspective.

Towards this goal, we introduce the problem of risk assessment from an agent-centric point of view.
That is, given the observed past and current behavior of each agent in a video, we tackle the problem by
answering two key questions centered around each agent.
First, will the agent encounter an accident in the near future? This corresponds to the task of \emph{accident anticipation}, where we would like to predict an accident before it occurs. Second, in which region in the environment might the accident take place? This corresponds to the task of \emph{risky region localization}, where we would like to spatially localize the regions in the scene that might be involved in a future accident.

\begin{figure}[t]
\begin{center}
\includegraphics[width=0.48\textwidth]{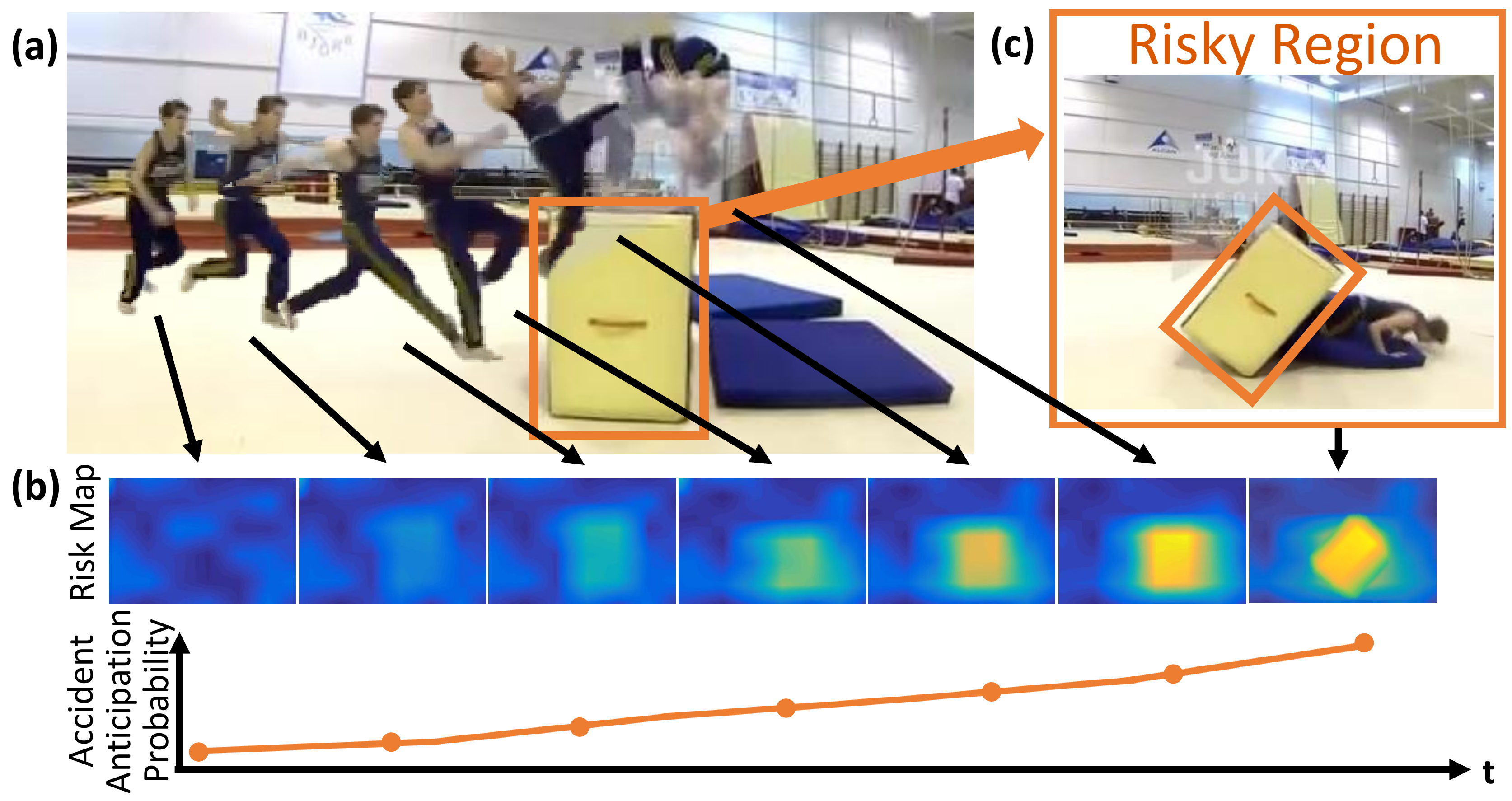}
\end{center}
\vspace{-2mm}
\caption{\small Illustration of risk assessment. (a) we show an image overlaid with a human agent from different frames and label the risky region before an accident occurs in an orange box. (b) risk map for environment and accident anticipation probability through time $t$. (c) risky region (orange box) at the instant when an accident occurs.}\label{fig:teaser}
\vspace{-6mm}
\end{figure}


We face two key but difficult challenges in this problem.
First, note that similar visual appearances will frequently correspond to vastly different levels of risk, as risk is dependent
on context and interactions between the agent and the environment. Therefore, we must explicitly consider appearances and spatial
relations between agents and regions in the scene.
A second challenge is that of capturing long-term temporal dependencies and causalities that underlie risk events. This could be tackled by explicitly forecasting relationships between the agent and the environment.






Some early attempts focus on assessing risk related to the environment \cite{Valenzuela2013} or correlating the statistical occurrence of activities to static scenes \cite{CityCrime}.
Instead, we aim at assessing risk explicitly triggered by the actions of an agent and its interaction with the environment, by anticipating accidents and localizing risky regions in the scene.
The task of accident anticipation is related to early activity recognition \cite{Hoai-DelaTorre-CVPR12,Ryoo11}, and event anticipation \cite{JainICRA16,koppula2016anticipating}. However, these are primarily categorization approaches that discriminate actions into separate semantic classes. In our case, we are not as interested in the semantic categories of the actions, but in reasoning about the probability of an accident in the near future.
Risky region localization has less precursors in the vision literature. The closest is work on human-object interaction from an action recognition perspective,
but these methods usually model object categories explicitly and their correlation to action classes \cite{yao2010modeling,koppula2016anticipating}.


We introduce a novel model for agent-centric risk assessment. 
Our model encodes the behavior of an agent into a distributed representation using a Recurrent Neural Network (RNN).
Given the agent representation, we introduce a novel dynamic parameter predictor inspired by Noh et al.~\cite{noh2015dppnet}
to measure the riskiness of each region with respect to the agent. The parameters efficiently consider relative spatial relations and coupled appearances between the agent and the region.
Next, our model takes the agent representation and appearance of the risky regions as the input of another temporal-level RNN for accident anticipation.
Moreover, the hidden representation of the temporal-level RNN is used to imagine and simulate the future trajectory of the agent. The future trajectory can be used as new inputs to our model so that we can assess the risk in long-term.
Our main contributions are:
\emph{(i)} We utilize the dynamic parameter layer to efficiently model the relative spatial relation and coupled appearance between agent and region.
\emph{(ii)} We use the generative property of RNN to self-train it to encode the behavior of the agent as well as generate (i.e., imagine) its future trajectory.
\emph{(iii)} The imagined future trajectory becomes new inputs to our model to assess risk in a longer term.
\emph{(iv)} To the best of our knowledge, the new Epic Fail (EF) video dataset is the first agent-centric risk assessment dataset for computer vision research.

\cutsectionup\section{Related Work}\label{sec.RW}\cutsectiondown
We give an overview of related work on risk assessment from visual observations, early event recognition and anticipation, as well as parameter prediction for deep networks.

Risk assessment given visual observation has not been widely explored.
Valenzuela \etal\cite{Valenzuela2013} propose to assess landslide risk from topographic images. Since landslide is caused by intense rain in localities where there was an unplanned occupation of slopes of hills and mountains, detecting these slopes in topographic images helps us to predict the risk of landslide.
Arietta \etal\cite{CityCrime} propose to use street-level images to predict the crime rate (risk of crime) at each geographic location. Koshla \etal\cite{CVPR14_Khosla} predict crime rates in an area without real-time criminal activity information, by correlating the appearance of a scene to properties such distance to public places, businesses, etc. However, these approaches assess risk caused either by the environment or by priors on social activities, whereas we focus on assessing risk explicitly triggered by the observed actions of an agent and its interactions with the environment.

Risk assessment is related to predicting the possibility of catastrophic events occurring in the future.
In early activity recognition, the focus is to predict activities before they are completed, such as recognizing a smile as early as the corners of the mouth curve up.
For example, Ryoo~\cite{Ryoo11} introduces a probability model for early activity prediction;
Hoai \etal\cite{Hoai-DelaTorre-CVPR12} propose a max-margin model to handle partial observation;
and Lan \etal\cite{Lan2014} propose the hierarchical movemes representation for predicting future activities.
In activity anticipation, the goal is to predict events even before they occur.
For instance, Jain \etal\cite{JainICRA16} propose to fuse multiple sensors to anticipate the actions of a driver;
Chan \etal\cite{ACCV_accident} introduce a dynamic soft-attention-based RNN to anticipate accidents on the road from dashcam videos;
and Vondrick \etal\cite{Vondrick_2016_CVPR} propose to learn temporal knowledge from unlabeled videos for anticipation.
However, these focus on activity categories and do not study risk assessment of objects and regions in the video.

Anticipation has been applied in tasks other than event anticipation.
Kitani \etal\cite{Kitani_2012_7250} propose to forecast human trajectory by surrounding physical environment (e.g., road, pavement, etc.) and show that the forecasted trajectory can be used to improve object tracking accuracy.
Walker et al.~\cite{walker2016uncertain} propose to forecast dense pixel trajectories from a static image.
Yuen and Torralba~\cite{Yuen:2010} propose to predict motion from still images.
Julian \etal\cite{walker2014patch} propose a novel visual appearance prediction method based on mid-level visual elements with temporal modeling methods.
Event anticipation is also popular in the robotic community.
Wang \etal\cite{WangDBVSP2012} propose a latent variable model for inferring human intentions.
Koppula and Saxena~\cite{koppula2016anticipating} address
the problem by observing RGB-D data, and apply their method to assist humans in daily tasks.
Finally, human activity anticipation can also improve
human-robot collaboration \cite{Koppula2016,BerensonIROS13}.

Parameter prediction in deep networks is a relatively new idea.
Ba \etal\cite{BaICCV15} propose a zero-shot classifier for unseen classes by predicting the parameters of a classifier using text information.
Noh \etal\cite{noh2015dppnet} propose to dynamically predict the parameters for image question answering depending on the given textual question. Inspired by \cite{noh2015dppnet}, we introduce a novel dynamic parameter predictor layer for estimating spatial riskiness depending on the agent behavior.



\begin{figure*}[!t]
\begin{center}
\includegraphics[width=0.85\textwidth]{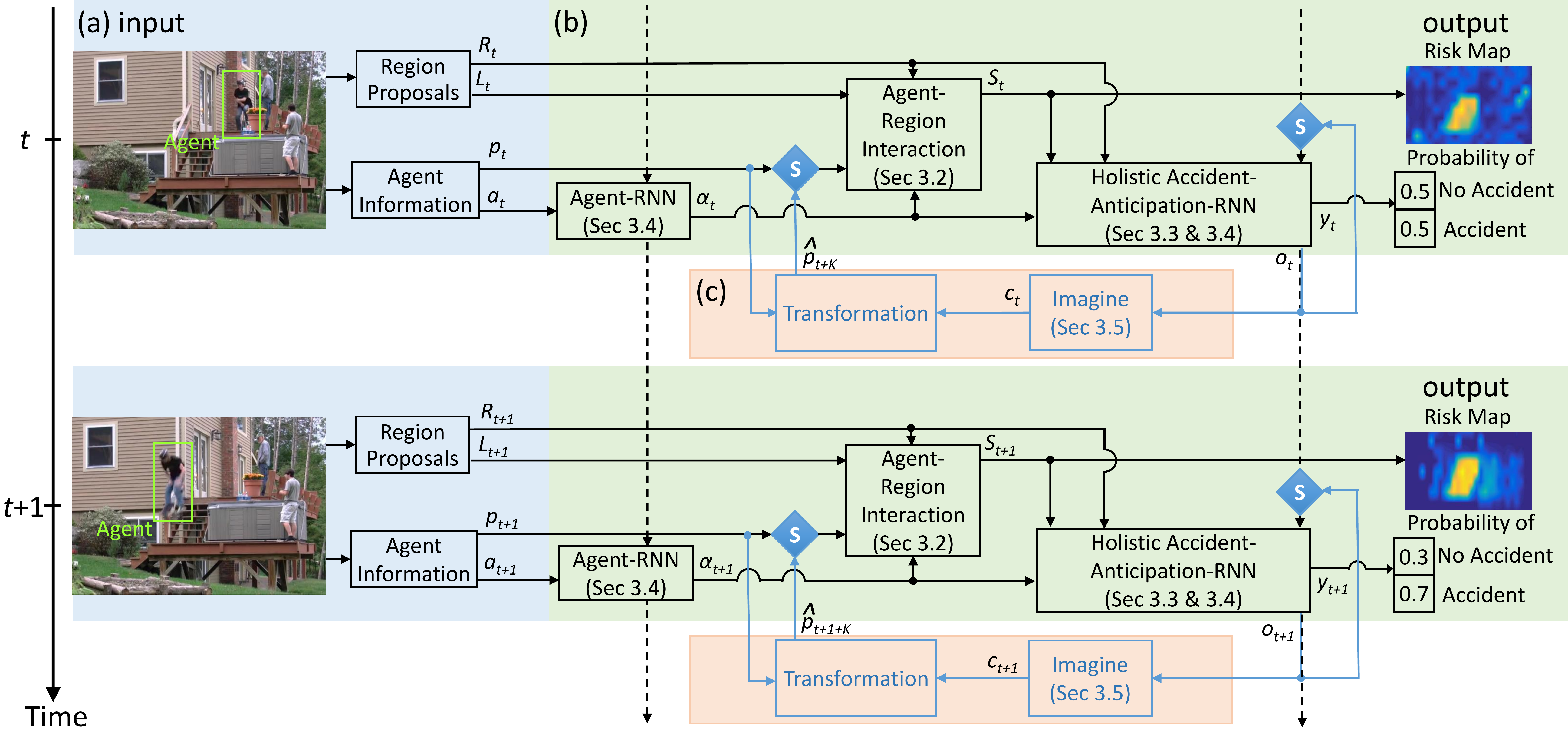}
\end{center}
\vspace{-4mm}
    	\caption{\small Illustration of our method. Panel (a) shows the pre-process to obtain appearance and location information for both agent $(a,p)$ and regions $(R,L)$.
        Panel (b) shows all the components (Sec.3,2, 3.3, and 3.4) in our model to predict riskiness of all regions $S$ and anticipated accident probability $y$. Acc. denotes accident.
        Panel (c) illustrates how the imagined agent location $\hat{p}_{t+K}$ triggers our model to reassess risk ($S,y$). In panel (b), a diamond shape node denotes a switch. It is used to control our model to imagine or take the observation. In panel (c), the transformation block corresponds to Eq.~\ref{Eq.b}. All dash arrows represent information across frames. Note that the anticipated accident probability increases from 0.5 at frame t to 0.7 at frame t+1.
        }\label{fig:pipeline}
        \vspace{-6mm}
    	\end{figure*}

\cutsectionup\section{Agent-centric Risk Assessment}\label{sec.Tech}\cutsectiondown

We now define the task of agent-centric risk assessment and present our model.
Given a video frame at time $t$, we observe information about the agent and multiple regions.
We assume we have access to the appearance vector $a_t$ and bounding box location $p_t = [x_t, y_t, w_t, h_t]$ of the agent.
We also capture information about a set of $N$ candidate risky regions, $R_t = \{r_t^i\}_{i=0}^{N}$ and $L_t = \{l_t^i\}_{i=0}^{N}$, where $r_t^i$ is the appearance and $l_t^i$ the location of region $i$.
When we observe a video sequence from $t=0$ to the current frame $\hat{t}$, our accumulated agent information is $\{(a_t,p_t)\}_{t=0}^{\hat{t}}$ and our accumulated region information is $\{(R_t,L_t)\}_{t=0}^{\hat{t}}$.
The goal is to predict two outputs corresponding to the tasks of accident anticipation and risky region localization.
The first is the accident anticipation probability $y_{\hat{t}}\in \left[0, 1\right]$ at current frame $\hat{t}$. The second is the riskiness score of all candidate regions at current frame $\hat{t}$, $S_{\hat{t}} = \{s_{\hat{t}}^i\}_{i=0}^N$, where $s_{\hat{t}}^i\in \left[0, 1\right]$ is the risk probability for the $i$-th region. Next, we give an overview of how our model infers $y_{\hat{t}}$ and $S_{\hat{t}}$.



\cutsubsectionup
\subsection{Model Overview}
\cutsubsectiondown
Our model consists of three main components. The first is the agent-region interaction component. We propose to dynamically predict parameters to infer riskiness of a region $s$ depending on the behavior of the agent and relative location of the region concerning the agent's location.
The second is the Holistic Accident Anticipation Module incorporating information from both agent and risky regions to infer the accident anticipation probability.
Finally, the recurrent component with two Recurrent Neural Networks (RNNs). One RNN aggregates behavior of the agent, while the other aggregates the holistic accident anticipation information.
In the following, we describe each component in details.

\cutsubsectionup\subsection{Agent-Region Interaction Module}\label{subsec.IM}\cutsubsectiondown

\begin{figure}[t]
\begin{center}
\includegraphics[width=0.36\textwidth]{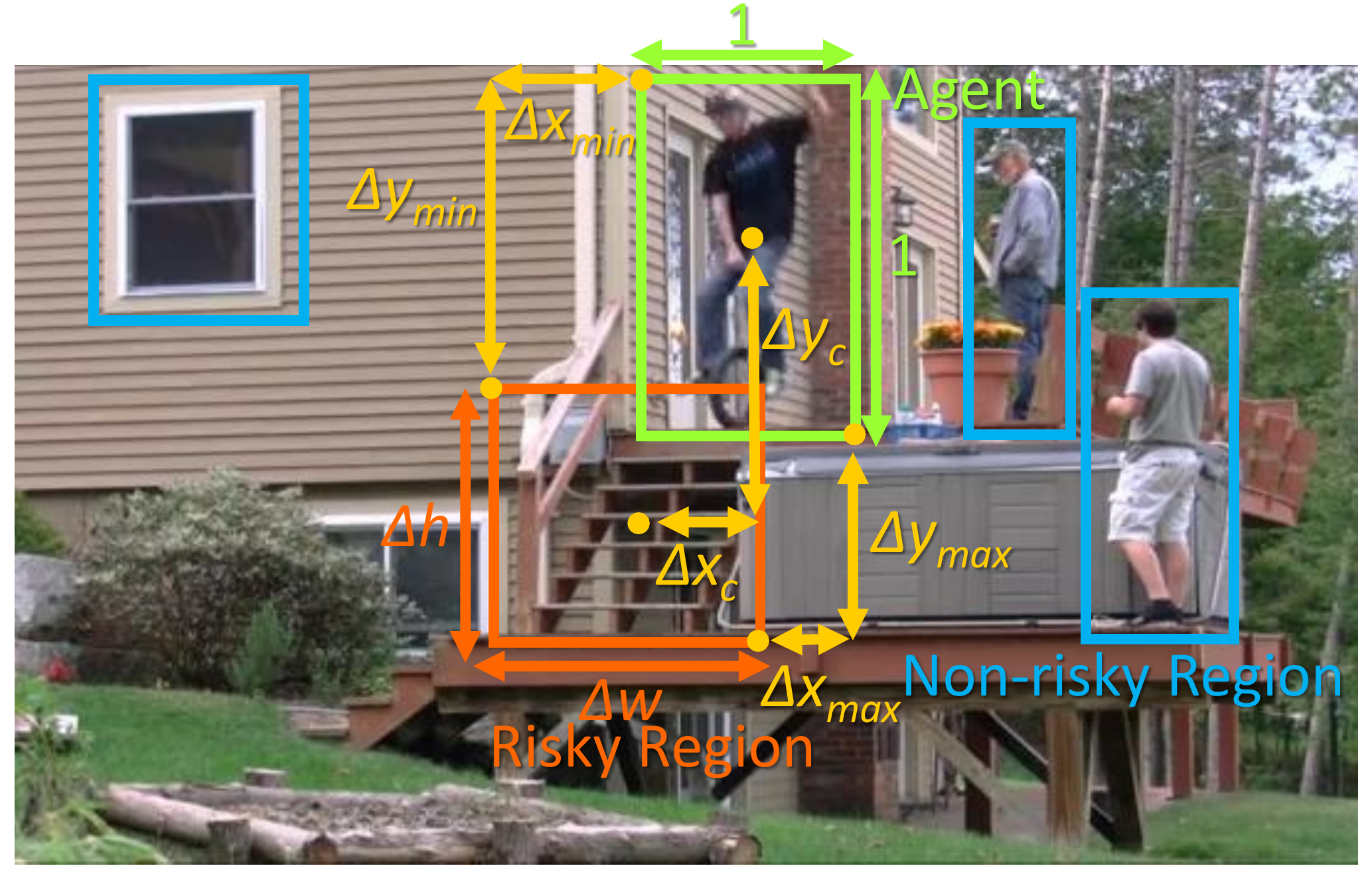}
\end{center}
\vspace{-4mm}
\caption{\hao{Relative configuration of all regions with respect to the agent. Risk assessment of all regions are illustrated with respect to the agent (green box). In our agent-centric perspective, the orange box indicates a risky region and the blue boxes indicate non-risky regions.} We normalize the horizontal and vertical axes separately such that the width and height of the agent are unit one. All 9 cues ($\Delta x_c, \Delta y_c, \Delta x_{min}, \Delta y_{min}, \Delta x_{max}, \Delta x_{max}, \Delta w, \Delta h$, IoU) in the configuration are visualized. 
}
\label{fig.RC}
\vspace{-6mm}
\end{figure}

The goal of this module is to infer the risk probability $S^i$ for each region in a frame.
Consider for example the unicycler agent in frame $t$ of Fig. \ref{fig:pipeline}(a).
Intuitively, the risk of region $i$ should be dependent on:
the appearance of the region $r^i$ to verify if the objects in a region are risky,  as a region covering the stairs;
the appearance of the agent $a$, as the stairs might be riskier for a unicycler than for a pedestrian;
and spatial relationship between the agent and the region $u^i$, as stairs close to the unicycler indicate more risk.
In this way, we write risk probability $s^i$ as:
\small\begin{equation}
s^i=g(w_r^T \cdot r^i)\in \left[0, 1\right], \label{Eq.RR}
\end{equation}\normalsize
where $g$ is a sigmoid to ensure valid probability estimates.
Note that this indicates that region riskiness only depends on region appearance $r^i$.
To encode dependencies on $a$ and $u^i$,
we propose to dynamically predict the parameter $w_r$:
\small\begin{equation}
w_r(a,u^i)=\sigma\left(W_f \cdot \begin{bmatrix} a & \sigma\left(W_u \cdot u^i\right) \end{bmatrix} ^T \right), \label{Eq.DP}
\end{equation}\normalsize
where $\sigma$ is a rectified linear unit (ReLU), and $W_f,W_u$ are the parameters of two fully connected layers.
We encode the agent-region spatial relationship $u^i$ with a $9$-dimensional vector that we compute from the agent bounding box $p$ and region bounding box $l^i$. Fig. \ref{fig.RC} illustrates 
the components of $u^i$ which concatenates:
the normalized relative position of region center $(\Delta x_{c},\Delta y_{c})$,
top-left corner $(\Delta x_{min},\Delta y_{min})$ and
bottom-right corner $(\Delta x_{max},\Delta y_{max})$; the region relative width $\Delta w$ and height $\Delta y$;
and Intersection over Union (IoU) of the agent box and region boxes.


\cutsubsectionup\subsection{Holistic Accident Anticipation Module}\label{subsec.aware}\cutsubsectiondown

The goal of this module is to produce an accident anticipation score $y$ for the current frame.
Intuitively, the probability of accident $y$ depends on:
the appearance of the agent $a$, as some agents might be more prone to accidents than others;
and the appearance $R$ and risk level $S$ of all regions in the scene, as some specific types of regions might lead to accidents more frequently than others.
We encapsulate this intuition by first building a holistic representation $q$, which we obtain by concatenating the agent appearance $a$ with the consolidated region information $\bar{r}$:
\small\begin{equation}
    q = \begin{bmatrix} a & \bar{r} \end{bmatrix}^T. \label{Eq.q}
\end{equation}\normalsize
We consolidate the region information by weighting  each region according to its inferred risk probability:
\small\begin{equation}
    \bar{r}=\phi(S,R)=\textstyle{\sum}_{i}s^i\cdot r^i.\label{Eq.RAt}
\end{equation}\normalsize
Note that $\bar{r}$ has the same dimension even when the number of regions varies at each frame.
The holistic representation $q$ is used to infer accident anticipation probability $y$,
\small\begin{eqnarray}
y=\softmax(W_y \cdot q)\in [0,1]^2, \label{Eq.y}
\end{eqnarray}\normalsize
where $W_y$ is the model parameter, and $y[0],y[1]$ denote the probability of non-accident and accident, respectively.



\cutsubsectionup
\subsection{Recurrent Temporal Memory for Anticipation}\label{subsec.mem}
\cutsubsectiondown
The model we described so far operates on a single frame and does not aggregate the knowledge of the sequence of past observations.
Intuitively, incorporating this sequence should help the model understand how the agent and regions move and how their relation with each other evolve in time.
To model these sequences, we introduce two RNNs to operate as memory components in our framework.

First, we aggregate the agent appearance and behavior information in the Agent-RNN ($RNN_A$), which takes $\{(a_t, p_t)\}_{t=0}^{\hat{t}}$ as inputs
and produces an encoding in its hidden vector $\alpha_{\hat{t}}$.
We propagate this information by incorporating $\alpha$ in Eq. \ref{Eq.DP} and Eq. \ref{Eq.q}, instead of the appearance information $a$.
So, Eq. \ref{Eq.DP} and Eq. \ref{Eq.q} can be rewritten as follows.
\small
\begin{align}
w_r(\alpha,u^i)&=\sigma\left(W_f \cdot \begin{bmatrix} \alpha & \sigma\left(W_u \cdot u^i\right) \end{bmatrix} ^T \right),\\
    q &= \begin{bmatrix} \alpha & \bar{r} \end{bmatrix}^T. \label{Eq.DP_q_RNN}
\end{align}
\normalsize

Second, we aggregate the environment risk information by modeling the sequence of holistic representations $q$. We achieve this by an Accident-Anticipation-RNN ($RNN_{AA}$), which takes $\{q_t\}_{t=0}^{\hat{t}}$ as input and produces an encoding in its hidden vector $o_{\hat{t}}$.
We propagate this information by incorporating $o$ in Eq. \ref{Eq.y}, instead of the direct use of $q$.

As a result, our model can predict the accident probability $y_{\hat{t}}$ and region risk scores $S_{\hat{t}}$ as a function of the observations from $t=0$ to $\hat{t}$.
In practice, we use LSTM cells \cite{LSTM} to better handle temporal dependencies.

\cutsubsectionup
\subsection{Imagining Future Risk}\label{subsec.Imagining}
\cutsubsectiondown
One interesting capability for humans is to assess risk by imagining future situations. In the case of Fig. \ref{fig:pipeline}, we can imagine the agent moving towards the stairs, which may result in an accident in the near future.
We are interested in encoding such \emph{imagination} capability to our model to better anticipate accidents and predict region risk.
With the formulation so far, we have a model that can predict the probability of an accident happening in the near feature
$t_f > \hat{t}$ from past observations $t=0$ to $\hat{t}$.
We include a mechanism in our model that simulates or imagines the future trajectory and location of the agent $K$ frames into the future, which we denote as $\hat{p}_{\hat{t}+K}$.
The idea is that once the model predicts the location of the agent in the future, we can ultimately produce new risk scores for all regions $\hat{s}$ and a new accident anticipation probability $\hat{y}$.

In practice, we use the holistic representation $o_{\hat{t}}$ to infer a 4-dimensional transformation
$\textbf{c} = [c_x, c_y, c_w, c_h]$ that converts the agent location $p_{\hat{t}}$ to the imagined location $\hat{p}_{\hat{t}+K}$:
\small\begin{align}
    \textbf{c} &= W_c \cdot  o_{\hat{t}} ,\label{Eq.b}\\
    \hat{p}_{\hat{t}+K} & = \begin{bmatrix} c_x \cdot w_{\hat{t}} + x_{\hat{t}} &
                                            c_y \cdot h_{\hat{t}} + y_{\hat{t}} &
                                            e^{c_w} \cdot w_{\hat{t}} &
                                            e^{c_h} \cdot h_{\hat{t}}\end{bmatrix}.\label{eq.m}
\end{align}\normalsize
We train the parameters $W_c$ with ground truth transformations $\textbf{c}^{*}$
that map ground truth locations $p_{\hat{t}}$ and $p_{\hat{t}+K}$.

Once the model imagines the location of the agent $\hat{p}_{\hat{t}+K}$, we can update the agent-region relationships to $\hat{u}_{\hat{t}+K}$ by recomputing these features using the imagined location.
Similarly, we can produce new $\hat{w}_r$, $\hat{S}_{\hat{t}+K}$, $\hat{\bar{r}}_{\hat{t}+K}$, $\hat{q}_{\hat{t}+K}$, $\hat{o}_{\hat{t}+K}$ and finally a new $\hat{y}_{\hat{t}+K}$.
Note that $\hat{y}_{\hat{t}+K}$ corresponds to the accident anticipation probability that the model produces from the observations at $t=0\ldots\hat{t}$ and one step of imagining the future position of the agent at time $\hat{t}+K$.
In other words, by using this imagination mechanism, the model is able to assess risk without observing any new information.
More importantly, the same process can be applied multiple times to imagine further into the future.
That is, we can obtain $\hat{y}_{\hat{t}+nK}$ by recursively estimating $\hat{p}_{\hat{t}+nK}$ from $\hat{p}_{\hat{t}+(n-1)K}$ and repeating the process outlined above. See more details in Appendix \ref{C_imagining_layer}.

\noindent\textbf{Final prediction.}
Finally, we estimate risk ($y_{\hat{t}}^F,S_{\hat{t}}^F$) by fusing the current risk with the imagined risk as follows,
\begin{eqnarray}
y_t^F=\sum_{n=0}^I\lambda_n \hat{y}_{\hat{t}+nK} \textrm{, and }
S_{t}^F=\sum_{n=0}^I\lambda_n \hat{S}_{\hat{t}+nK},\label{eq.imginf}
\end{eqnarray}
where, $\lambda_n$ are hyper-parameters, and with slight abuse of notation, we use $\hat{y}_{\hat{t}}$ as $y_{\hat{t}}$ and $\hat{S}_{\hat{t}}$ as $S_{\hat{t}}$.

\cutsubsectionup
\subsection{Multi-task Learning}\label{sec.loss}
\cutsubsectiondown

The goal of the learning process is to fit all the parameters in our model: $W_f, W_u, W_y, W_c$, and the parameters of our recurrent models $RNN_A$ and $RNN_{AA}$.
During training, we have access to a set of positive videos that depict accidents and a set of negative videos that depict normal non-accident events.
We assume each positive video depicts an accident at time $t=T$ and is annotated with the ground truth agent locations $p_0, \ldots, p_T$.
We also have access to the region bounding boxes $\rho_0, \ldots,\rho_T$ that encapsulate the part of the environment involved in the accident with the agent.

We fit the model parameters using these training examples by minimizing a loss function $\mathcal{L}$ over the multiple tasks that our model performs:
accident anticipation, risky region localization, and agent location imagination as follows:
\small\begin{equation}
    \mathcal{L} (Y, S, C)  = \mathcal{L}^A(Y) + \mathcal{L}^R(S) + \mathcal{L}^P(C).\label{Eq.loss}
\end{equation}\normalsize

\noindent\textbf{Accident Anticipation.}
We follow \cite{ACCV_accident} to use regular cross-entropy for non-accident sequences and exponential cross-entropy loss for accident sequences.
The exponential loss emphasizes on predictions for times $t$ that are closer to $T$. 
\small\begin{align}
    L^A(y_t) = & \begin{cases} -\log(y_t[0]) & \mbox{for non-accident} \\
                               -e^{-(T-t)}\log(y_t[1]) & \mbox{for accident}\label{Eq.pos_neg_accident}
\end{cases}\\
\mathcal{L}^A(Y) = & {\textstyle\sum}_{t=0}^T L^A(y_t),\label{Eq.yloss}
\end{align}\normalsize
where $Y=\{y_t\}_{t=0}^T$. 

\noindent\textbf{Risky Region Localization.}
Since $s$ is the output of a sigmoid function, we use sigmoid cross entropy loss for risky region localization as follows,
\small\begin{align}
    L^R(y) & = \begin{cases} -\log(1-s) & \mbox{for non-risky region} \\ -\log(s) & \mbox{for risky region}\label{Eq.pos_neg_region}
\end{cases} \\
\mathcal{L}^R(\mathcal{S}) & ={\textstyle\sum}_{t=0}^T {\textstyle\sum}_{i=0}^N L^R(s_t^i),\label{Eq.rloss}
\end{align}\normalsize
where $\mathcal{S}=\{S_t\}_{t=0}^T$, and the $i$-th region is a risky region if the IoU between region location $l^i$ and any ground truth box in $\rho_t$ is over $0.4$.

\noindent\textbf{Agent location imagination.}
Inspired by \cite{ren2015faster}, we employ a smooth $\ell_{1}$ loss $L^P(\textbf{c}_t,\textbf{c}^*_t)$ for agent location imagination:
\small\begin{equation}
    \mathcal{L}^P(C) = {\textstyle\sum}_{t=0}^T L^P(\textbf{c}_t,\textbf{c}^{*}_t),\label{Eq.closs}
\end{equation}\normalsize
where $C=\{\textbf{c}_t\}_{t=0}^T$
and $\textbf{c}^*$ is ground truth transformation.

\noindent\textbf{Anticipation and Localization with Imagination loss.}
As described in Sec.~\ref{subsec.Imagining}, once we imagine the future agent location $\hat{p}$, we can
update $\hat{y}$ and $\hat{s}$ iteratively. We incorporate these estimates by rewriting Eq. \ref{Eq.loss} as:
\small\begin{equation}
    \mathcal{L}_{I} (Y, S, C)  =  \mathcal{L}^P(C) +  {\textstyle\sum}_{n=0}^I \lambda_n \left( \mathcal{L}^A(Y^{n}) + \mathcal{L}^R(S^n) \right).\label{Eq.iterative_loss}
\end{equation}\normalsize
where $Y^n$ and $S^n$ are the predictions at the $n$-th imagination iteration, $I$ is the number of imagination steps, and $\lambda_n$ are the same hyper-parameters in Eq.~\ref{eq.imginf}. Because our model is fully differentiable, the model can be end-to-end training.

\noindent\textbf{Training with noisy agent info.}
Training and evaluating with ground truth agents does not reflect the challenges of handling noisy agent information as in real world.
Hence, we apply online tracking-by-detection (TD) to obtain candidate agent tracks (see Appendix \ref{A_online}
for detail). At training, we utilize both candidate agent and ground truth tracks. The ground truth accident labels are shared among all tracks. At testing, we obtain candidate agent tracks using the same approach. For each candidate agent track, we apply our method to obtain per frame accident anticipation probability. At each frame, we take the maximum probability as the video-level anticipation probability. Then, we take the estimated riskiness of regions from the agent with maximum accident probability as the final per frame riskiness.

\noindent\textbf{\hao{Summary of supervision.}} \hao{In training, temporal location of the accident within the video and bounding boxes of risky region at each frame in the positive examples. In testing, this information is only used for performance evaluation.}

\cutsubsectionup
\subsection{Implementation Details}
\cutsubsectiondown

We set $I=1$, $K=5$, $\lambda_0 = 0.6$ and $\lambda_1=0.4$.
This are set empirically
without heavily tuning.
At each frame, we use Faster R-CNN \cite{ren2015faster} to propose $300$ candidate risky regions. We find this setting to be effective, since the average recall at $0.4$ IOU is $79.5$\%, $74.9\%$ on Epic Fail dataset and Street Accident dataset, respectively. For each candidate risky region, we extract $pool 5$ feature and utilize Global Average Pooling \cite{lin2013network} to obtain a one dimensional representation. For an agent, we extract $fc7$ feature as the representation. All the feature extractor is using VGG16 model~\cite{Simonyan14c}.
We use Adam \cite{KingmaB14} as optimizer with default hyperparameters and 0.0001 learning rate and set batch size by $5$.
The model selection is done by early stopping \cite{haykin1994neural}.


\cutsectionup
\section{Dataset}\label{sec.Dataset}
\cutsectiondown
In order to evaluate our method, 
We collect a large-scale Epic Fail (EF) dataset consisting of user-generated videos, where a large portion of them involves epic ``human" accidents such as the parkour failure in Fig.~\ref{fig:teaser}. We also evaluate on the latest Street Accident (SA) dataset~\cite{ACCV_accident}, where both humans and/or vehicles can involve in accidents. We further describe each dataset in detail.




\begin{figure}[t!]
\begin{center}
\includegraphics[width=0.43\textwidth]{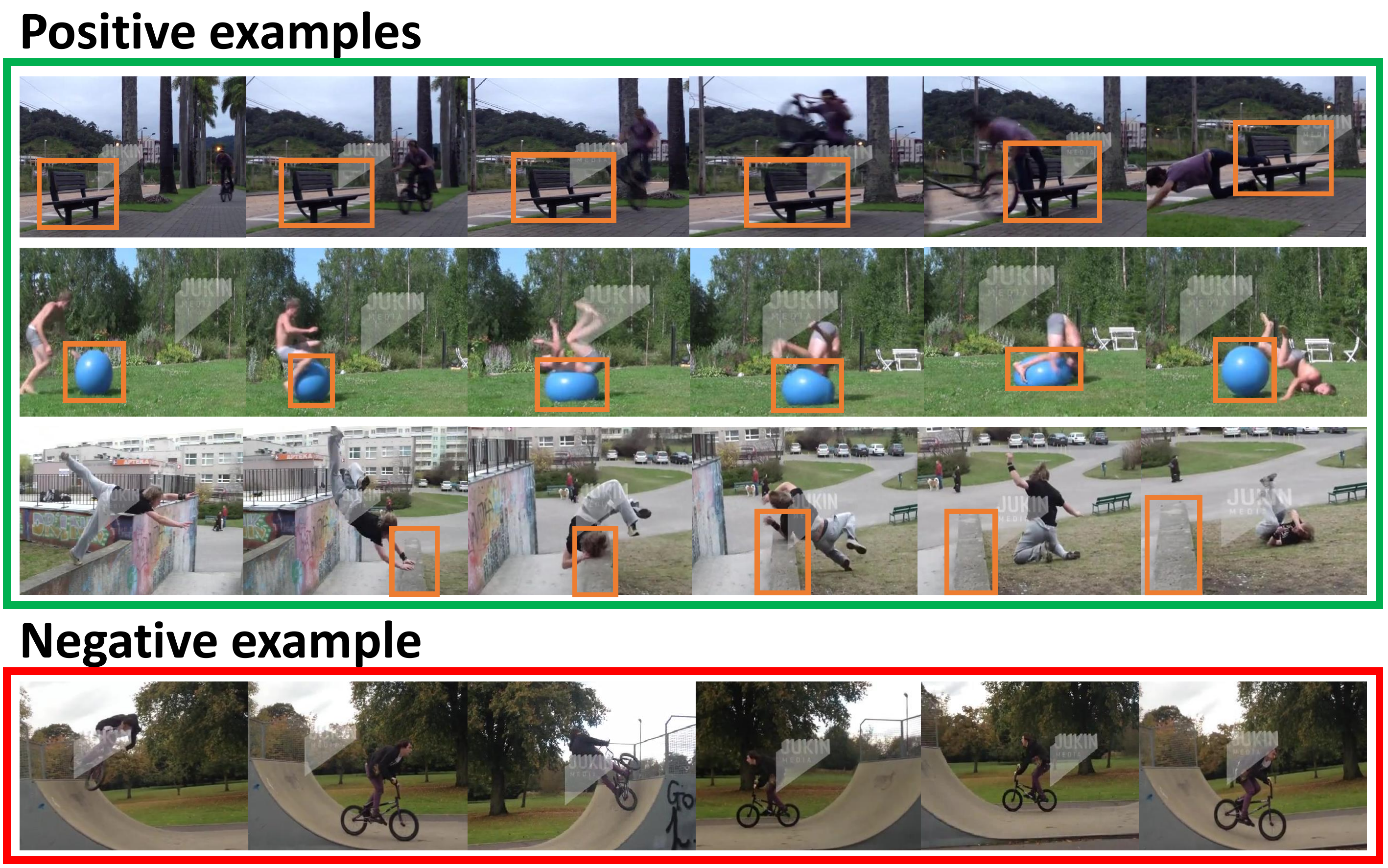}
\end{center}
\vspace{-4mm}
\caption{Examples in EF dataset. In each row, we show sampled frames from a video. For positive videos, we also show the annotated risky regions (orange boxes).}
\vspace{-6mm}
\end{figure}

\cutsubsectionup
\subsection{Epic Fail (EF) dataset}\label{sec.EAD}
\cutsubsectiondown

The raw videos in EF dataset are harvested from YouTube channels and Zeng et al.~\cite{ZengECCV16,zeng2017leveraging}. 
To build our new EF dataset, we first manually identify the time when an accident occurs in a subset of raw videos. Then, we sample short videos of 3-4 seconds from the subset. In total, we sample 3K videos and slip them into 2K training and 1K testing videos. In the training set, there are 1K positive videos and 1K negative videos. In the testing set, there are 609 positive videos and 391 negative videos.  
For positive videos, we ensure accident happens at the end of each video. For negative videos, we ensure no sign of accident appears. Note the types of accidents in our dataset is very diverse which include all kinds of skateboarder failure, skier failure, parkour failure, etc.

In order to train and evaluate the risk assessment performance, we ask users to annotate the dataset with the following ground truth labels.
Firstly, all videos (both positive and negative) are annotated with ground truth agent trajectory. Risky regions in all positive videos are also annotated, where we ask annotator to annotate the region causing the failure event.
The tool that users use for annotating the agent and risky region is an interactive annotation and segmentation tool called iSeg developed by \cite{schoningpixel}. The agents and the risky regions are annotated by bounding boxes.
Even with the help of the annotation tool, annotating bounding boxes are still time-consuming. Hence, training data is only annotated at every 15 frames. However, testing data is carefully annotated at every frame.
\hao{In this dataset, there are not too many cases with multiple risky regions because the dataset is collected from the user-generated videos. The user-generated videos typically have a main agent and an apparent region causing the accident.}
More detail and the data for EF dataset can be found in
Appendix \ref{EF_analysis} and our project page\footnote{Dataset can be downloaded at \url{http://aliensunmin.github.io/project/video-Forecasting/}}.

\cutsubsectionup
\subsection{Street Accident (SA) dataset}
\cutsubsectiondown

The SA dataset~\cite{ACCV_accident} is captured across six cities in Taiwan with high-quality dashcam (720p in resolution) and has diverse accidents occur in all videos consisting 100 frames. These accidents include 42.6\% motorbike hits car, 19.7\% car hits car, 15.6\% motorbike hits motorbike, and 20\% other types.
The SA dataset also provides the annotation about the time when an accident occurs and the trajectories of objects involved in the accident. The dataset consists of 596 positive examples containing the moment of accident at the last 10 frames, and 1137 negative examples containing no accident. In the SA dataset, it contains 1266 training videos (446 positive and 820 negative examples) and 467 testing videos (150 positive and 317 negative examples).
\hao{In this dataset, many cases have multiple risky regions because a street accident usually involves multiple vehicles.}



\cutsectionup
\section{Experiments}\label{sec.Exp}
\cutsectiondown

We first describe the baseline methods and variants of our method. Then, we define the evaluation metrics.
Finally, we show that our method achieves the best performance in both accident anticipation and risky region localization on both EF and SA datasets.




\noindent\textbf{Baselines.}
We compare the following state-of-the-art methods with our method.

\noindent\emph{- DSA:} Dynamic Soft-Attention \cite{ACCV_accident}.

\noindent \emph{- SP:} Social Pooling \cite{alahisocial}.
In the agent-centric representation, we apply SP~\cite{alahisocial} to pool the nearby regions information ($r,l$). The agent information ($a,p$) and the SP pooled feature are concatenated and fed into an LSTM for anticipating accident probability at each frame.



\noindent \emph{- R*CNN} \cite{gkioxari2015contextual}.
We extend the Contextual action classification method R*CNN~\cite{gkioxari2015contextual} for accident anticipation.
There are two extensions: (1) replacing classification loss with the same anticipation loss in Sec.~\ref{sec.loss}, and (2) removing the original IOU constraint for the model to observe all candidate risky regions.
Note that R*CNN uses hard-attention to select a region with maximum confidence, whereas our method uses soft-attention as in Eq.~\ref{Eq.RAt}.


\noindent \emph{- L-R*CNN}, an extended R*CNN to incorporate temporal modeling with LSTM.
We add a LSTM to aggregate information across time similar to the $RNN_{AA}$ in Sec.~\ref{subsec.mem}.


\noindent\textbf{Ablation studies.}
We also evaluate the following four variants of our Risk Assessment (RA) model involving adding memory or not and applying imagination or not. Note that w denotes ``with" and w/o denotes ``without".


\noindent\emph{- RA}.
w/o memory, w/o imagining.
This model observes a single frame without aggregating temporal information.

\noindent\emph{- RAI}.
w/o memory and w/ imagining.
We add imagining layer (Sec.~\ref{subsec.Imagining}) to the RA model.

\noindent\emph{- L-RA}.
w/ memory and w/o imagining.
We add LSTM cells (Sec.~\ref{subsec.mem}) to the RA model.

\noindent\emph{- L-RAI}.
w/ memory and w/ imagining.
This is our full model which can handle the temporal information and imagine the future.


\cutsubsectionup
\subsection{Evaluation Metrics}\label{subsec.evaluation}
\cutsubsectiondown

For accident anticipation, we are interested in not only the precision v.s. recall, but also the first time $\hat{t}$ when the anticipation probability is above a threshold $\gamma$. Let's assume the accident occurs at time $T$. We follow~\cite{ACCV_accident} and define Time-to-Accident (TTA) as $T-\hat{t}$.
Recall that given different $\gamma$, one can compute a precision and a recall. Similarly, we can compute TTA for each recalled positive video. This implies one can plot a TTA v.s. recall. We propose to report the ``average TTA" across different recall to summarize the TTA (referred to as ATTA). If the ATTA value is higher, the model can anticipate the accident earlier. We also report mean average precision (mAP) for all the videos. See detailed explanation of ATTA in Appendix \ref{B_ATTA}.


For the risky region estimation, we use the object detection metric~\cite{Everingham10} with $IOU\geq 0.4$ as the positive detection criteria~\cite{ACCV_accident}. This is because annotating ground truth boxes in videos is very time-consuming. As a result, the quality of ground truth boxes on both EF and SA dataset is slightly worse than other object detection dataset.
\hao{Note that each frame might contain more than one risky region. Moreover, risky region in a positive clip could appear and disappear due to occlusion or camera motion. Hence, the evaluation of risky regions is conducted per frame.}

\begin{table}[t!]
\centering
\small
\resizebox{0.45\textwidth}{!}{
\begin{tabular}{|c|c|c|c|c|}
\hline
Dataset     & \multicolumn{2}{c|}{EF} & \multicolumn{2}{c|}{SA} \\ \hline \hline
w/o memory & mAP (\%) & ATTA (s)   & mAP (\%)   & ATTA (s)   \\ \hline
R*CNN       & 68.6     & 2.47      & 40.7        & 2.64        \\ \hline
RA          & 72.2     & 2.10      & 47.8        & 2.55        \\ \hline
RAI         & 72.4     & 2.13      & 48.8        & 2.62        \\ \hline \hline
w memory & mAP (\%) & ATTA (s)   & mAP (\%)   & ATTA (s)   \\ \hline
DSA         & 45.7     & 1.16    &  48.1      &  1.34           \\ \hline
SP          & 40.5     & 0.88    &  47.3      &  1.66           \\ \hline
L-R*CNN     & \textit{69.6}     & \textit{2.54}    &  \textit{37.4}      &  \textit{3.13}           \\ \hline
L-RA        & 74.2     & 1.84    &  49.1      &  \textbf{3.04}           \\ \hline
L-RAI       & \textbf{75.1}     & \textbf{2.23}    &  \textbf{51.4}      &  3.01          \\ \hline
\end{tabular}
}
\vspace{1mm}
\caption{Quantitative results of accident anticipation. We evaluate accident anticipation by estimating mean average precision and average time-to-accident (ATTA) metrics.
Bold-fonts indicate our best performance.
Italics-fonts indicate best baseline performance.}
\label{table.accident}
\vspace{-4mm}
\end{table}

\cutsubsectionup
\subsection{Accident Anticipation}\label{sec.AAExp}
\cutsubsectiondown
\hao{
Quantitative results using both mAP and ATTA are shown in Table.~\ref{table.accident}.
For our model, adding memory improves mAP as well as ATTA on both datasets in general (i.e., L-RA outperforms RA and L-RAI outperforms RAI, except L-RA is worse than RA in ATTA on EF dataset.)
Imagining future risk effectively improves both evaluation metrics on both datasets (i.e., RAI outperforms RA and L-RAI outperforms L-RA)
On the other hand, RA/L-RA outperforms R*CNN/L-R*CNN significantly in mAP. This suggests that our soft-attention using dynamic parameter prediction outperforms hard-attention.
Although L-R*CNN outperforms our method in ATTA on both datasets, this earlier anticipation comes with significant more false alarms since there is a significant $\sim 5\%$ drop in anticipation mAP.
L-RA also outperforms both DSA and SP. This suggests that our dynamic parameter prediction layer is more effective than social pooling and dynamic soft-attention. Note that DSA and SP do not support risky region localization.
}

\begin{table}
\centering
\small
\resizebox{0.3\textwidth}{!}{
\begin{tabular}{|c|c|c|}
\hline
Dataset     & EF       & SA       \\ \hline \hline
w/o memory & mAP (\%) & mAP (\%) \\ \hline
R*CNN       & 3.47     & 34.7         \\ \hline
RA          & 12.3     & 40.1         \\ \hline
RAI         & 14.1     & 43.1        \\ \hline \hline
w memory & mAP (\%) & mAP (\%) \\ \hline 
L-R*CNN     & 3.5     &  35.6        \\ \hline
L-RA        & 14.0     &  43.8        \\ \hline
L-RAI       & \textbf{15.1}     &  \textbf{45.4}        \\ \hline \hline
Oracle      & 75.7     &  92.8        \\ \hline
\end{tabular}
}
\vspace{2mm}
\caption{Quantitative results of risky region estimation. We evaluate risky region using traditional object detection metric and compute mean average precision over the entire testing set. }
\label{table.risk_result}
\vspace{-4mm}
\end{table}

\cutsubsectionup
\subsection{Risky Region Localization}\label{sec.RLExp}
\cutsubsectiondown
\hao{
Quantitative results using mAP of risky region localization are shown in Table.~\ref{table.risk_result}. Note that the mAP cannot be $100\%$, since our results depend on the detections performed by Faster R-CNN for proposing candidate risky regions. Hence, we report the oracle performance in the last row which is achieved by assuming all candidate regions are classified correctly. This serves as the upper-bound performance.
For our model, adding memory module improves mAP on both datasets (i.e., L-RA outperforms RA and L-RAI outperforms RAI).
Imagining future risk effectively improves mAP on both datasets (i.e., RAI outperforms RA and L-RAI outperforms RAI).
On the other hand, L-RA/RA significantly outperforms L-R*CNN/R*CNN. This suggests that our soft-attention using dynamic parameter prediction outperforms hard-attention.
}


\cutsubsectionup
\subsection{Qualitative Results}\label{sec.te}
\begin{figure*}[t]\cuttableup
\begin{center}
\includegraphics[width=0.92\textwidth]{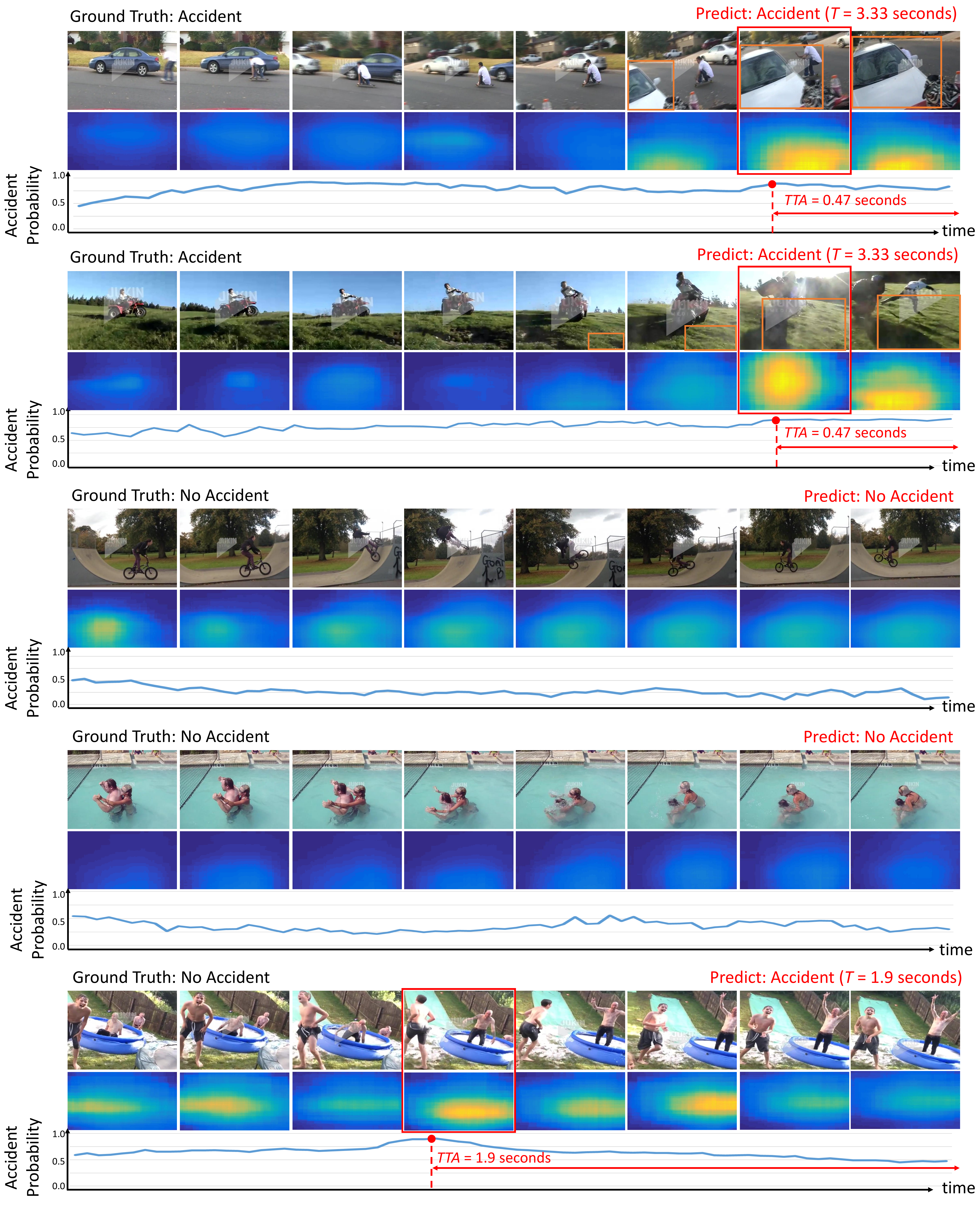}
\end{center}
\vspace{-2mm}
\caption{Qualitative Results. We set $0.9$ as the threshold of triggering accident anticipation to show the qualitative results. In each example, we show a typical example with accident anticipation probability (bottom row), the heat map (yellow for high risk and blue for low risk) for the risky regions (middle row) and ground truth risky region (orange box in top row). For risky heat map, we average risky confidences of covering boxes for each pixel and draw the map by using Matlab \cite{MATLAB:2010} imagesc tool. More detail for drawing heat map can be found in
Appendix \ref{E_risk_map}
. The first and second one are positive examples. The third and fourth one are negative examples. The last one is a failure case, where the model misunderstands risky regions so that it has higher accident anticipation probability at first. However, after long-term observation, the model correct the anticipation probability.}
\label{fig.vis}
\end{figure*}

\cutsubsectiondown
We show qualitative results of accident anticipation and risky region localization in Fig.~\ref{fig.vis}.
From the positive and negative examples, our method shows great ability to differentiate them. These examples also demonstrate the ability to localize risky regions of different categories (e.g., car and bars). In the failure example, our system identifies the inflatable pool as potential risky, which is fairly reasonable. See more qualitative results in Appendix \ref{F_qualitative}.

\vspace{-2mm}
\cutsectionup
\section{Conclusion}\label{sec.Con}
\cutsectiondown

We introduce new risk assessment tasks including (1) accident anticipation and (2) risky region localization. To tackle these tasks, we propose a novel model with two main innovations: (1) dynamic parameter prediction to capture the relative spatial relation and appearance-wise coupling between agent and risky regions.
Our proposed method outperforms baselines methods significantly on both accident anticipation and risky region estimation.
In the future, we plan to extend our imagining layer for the environment. We believe that stimulate both agent and environment in the future simultaneously would enhance the model and give a way to explain how does the model anticipate the accident.

\noindent\textbf{Acknowledgement.} We thank MOST 104-2221-E-007-089, MOST 106-2633-E-002-001, National Taiwan University (NTU-106R104045), NOVATEK Fellowship, MediaTek, and Panasonic for their support. We thank Chi-Wen Huang, and Yu-Chieh Cheng for their collaboration. We thank Alexandre Alahi, Zelun Luo, and Shyamal Buch for helpful comments and discussion.

{\small
\bibliographystyle{ieee}
\bibliography{egbib}
}
\newpage

\appendix
\cutsectionup
\section{Online tracking-by-detection for agent tracks}\label{A_online}
\cutsectiondown

We outline the training process in
Section 3.6 of the main paper. In particular, we describe the use of noisy
agent information for training, which comes from
adopting online tracking-by-detection (TD) to obtain candidate agent tracks.
Instead of using ground truth agent information during training,
the use of noisy tracks better reflects the challenge of handling noisy agent information during testing.

In detail, we first fine-tune faster R-CNN \cite{ren2015faster} on the training set of EF and SA datasets, which we use to propose $300$ regions for each frame. Then, we use the $30$ region proposals with top object scores as initial boxes. We further retrieve the top $50$ boxes from the $300$ region proposals in the next frame for each track by computing IoU scores. Then, the top one box is selected by the highest cosine similarity in feature space ($pooling\ 5$ with $Global\ average\ pooling$) to the selected box in current frame for each track.
When more than one track ends in highly overlapping boxes in the last frame, we only keep the one track from the group with the highest average object score.


During training, we randomly select one agent track from the set consisting of the annotated agent tracks and the automatically generated TD tracks in each epoch. For each video, we treat the selected track as a positive/negative agent for the positive/negative example. During testing, no annotations are available to the model, so we only use TD tracks. We follow the same procedure when evaluating the accident anticipation and risky region localization. For each video, we apply our method to obtain per frame accident anticipation probability for each candidate agent track. At each frame, we take the maximum probability as the video-level anticipation probability. Then, we take the estimated riskiness of regions from the agent with maximum accident probability as the final per frame riskiness.

\section{Average Time to Accident (ATTA)}
\label{B_ATTA}

In Section 5.1 of the main paper, we introduce Time-to-accident (TTA) and its averaged version (ATTA) to evaluate how early the model is able to predict an accident.
We also indicate the relation between TTA and the threshold $\gamma$ on accident anticipation probability. Here, we provide more details on the relation between TTA and $\gamma$ and deduce the definition of average-TTA (ATTA).

\begin{figure}[t!]
\begin{center}
\includegraphics[width=0.3\textwidth]{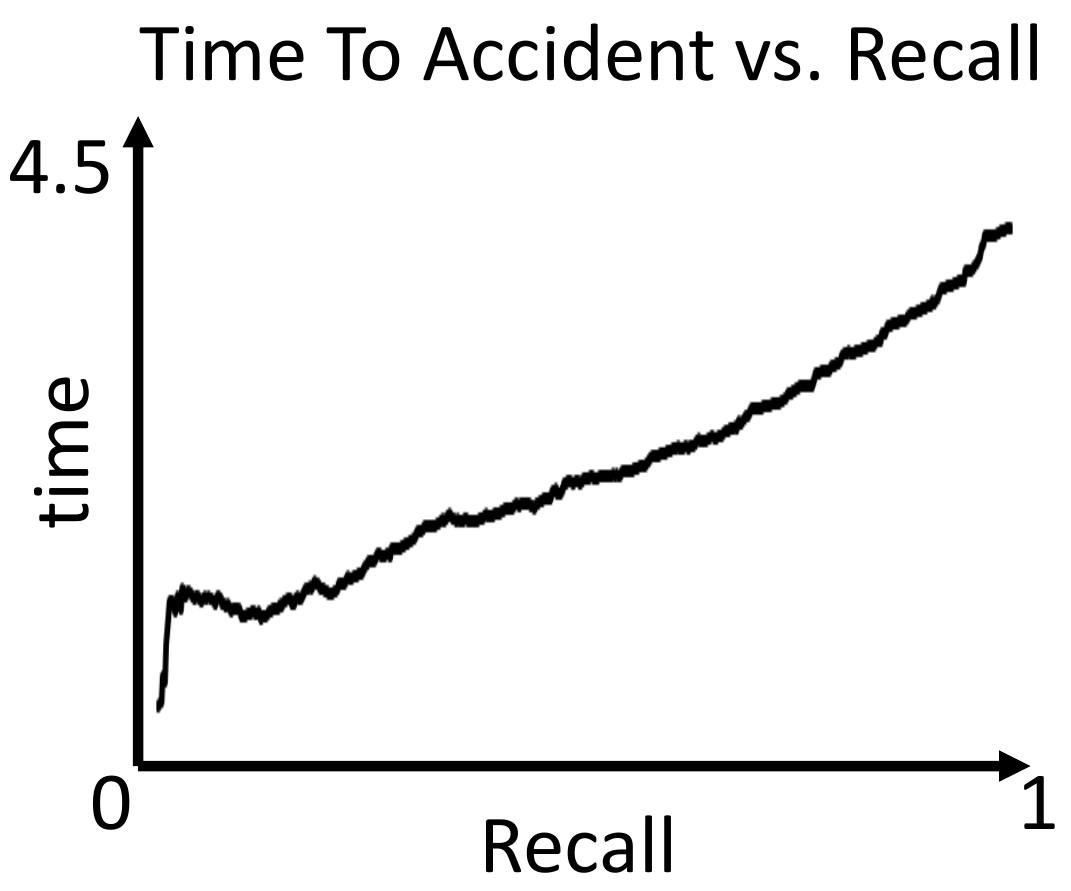}
\end{center}
\vspace{-2mm}
\caption{Time-to-Accidents vs. Recall. Higher recall results in higher TTA and vice versa. We simply average TTA across different recall, named as ATTA, as the summarization for TTA-Recall curve.}
\label{fig.TTA}
\end{figure}

Recall that for a given $\gamma$, we can compute one precision-recall operating point.
Similarly, we can compute TTA for each recalled positive video given a specific $\gamma$. Thus, we can plot a TTA against Recall as in Fig. \ref{fig.TTA}. From Fig. \ref{fig.TTA}, we find that higher recall (obtained with low $\gamma$) results in higher TTA and vice versa. To summarize all TTAs obtained with the various $\gamma$ settings (each producing different recall), we simply average TTA across recall. Therefore, if the ATTA value is higher, the model can anticipate the accident earlier.

\begin{table*}[]
\centering
\resizebox{0.97\textwidth}{!}{
\begin{tabular}{|c|c|c|c|c|c|c|c|c|c|}
\hline
Statistics & \multicolumn{4}{c|}{Positive}                    & \multicolumn{4}{c|}{Negative}                    & Total   \\ \hline
Items         & \#video & agent & risky region & annotation type & \#video & agent & risky region & annotation type & \#video \\ \hline
training set  & 1000    & v     & v            & every 15 frame  & 1000    & v     & x            & every 15 frame  & 2000    \\ \hline
testing set   & 609     & v     & v            & every frame     & 491     & v     & x            & every frame     & 1000    \\ \hline
\end{tabular}
}
\vspace{2mm}
\caption{VA dataset statistics}
\label{Table.stat}
\end{table*}

\section{Backpropagation for the imagining layer}
\label{C_imagining_layer}

Recall that we introduce the imagining layer in our system in the main paper. The imagining layer can simulate the future location of the agent according to current system status and feedforward the Agent-Region Interaction module to obtain a new accident anticipation probability and multiple risky region locations of the current frame. This simulation improves the performance on both accident anticipation and risky region localization according to Table 1 and 2 in the main paper, respectively. Here, we provide the detail of backpropagation of the imagining layer.

In the main paper, Eq. 17 is an overall loss function for accident anticipation, risky region localization and imagining of the location of an agent in the future. Taking Agent-RNN ($RNN_{A}$), a Recurrent Neural Network for appearance feature of the agent, as an example, the loss function for accident anticipation can be backpropagated through Eq. 5, Eq. 6, Eq. 7, Eq. 1 and Eq. 2 in the main paper for non-imagining step.


For imagining step, we take the next $t+k$ frame as our imagining target frame so that our system can imagine the location of the agent at time $t+k$.
As a result, the loss function for accident anticipation can be backpropagated through Eq. 5, Eq. 6, Eq. 7, Eq. 1, Eq. 2, Eq. 9 and Eq. 8 in the main paper to time equal $t$. Then we compute a weighted sum over the gradients produced by imagining step and the gradient produced by non-imagining step according to the loss function depicted in Eq. 17 in the main paper.

The gradients for other model parameters can also be computed in a similar manner. We use Adam \cite{KingmaB14} as our optimization approach. The combination of an imagining layer with recurrent neural network creates a novel idea for handling data with temporal information. We leave the deeper exploitation of it as a future work.


\section{Analysis for EF dataset}
\label{EF_analysis}

In Section 4.1 in the main paper, we introduce a new large-scale Epic Fail (EF) dataset. The raw videos in this dataset are harvested from YouTube channels. It consists of 3000 viral videos capturing various accidents.
In total, we have 1609 positive videos annotated as accident videos, while the rest are negative videos without accidents. In practice, we categorize each accident video into $7$ failure classes, each agent into $17$ object classes and each risky region into $63$ object classes. The statistics of the VA dataset and the details of all classes are presented in found in Table \ref{Table.stat} and Table \ref{table.class}, respectively. Although all these ground truth data are available, our method only requires annotations for the agent track, the risky region localization and the time that accident event happens for training. In testing time, our method only uses a raw image for each time as the input data.

\begin{table}[]
\centering
\begin{tabular}{|c|c|}
\hline
              & Classes                                                                              \\ \hline
Failure event & \begin{tabular}[c]{@{}c@{}}crash, fall, bump, hit, \\ turnover, hurt, burned\end{tabular}                                            \\ \hline
Agent         & \begin{tabular}[c]{@{}c@{}}bike, board, car, motorcycle, \\ motorcyclist, person, pushups device, \\ skateboarder, skier, sledge, swim ring, \\ toy, couch, tricycle, unicycle,  \\ air raft\end{tabular}                                                                           \\ \hline
Risky region  & \begin{tabular}[c]{@{}c@{}}ball, balloon, bar, barrier, \\ basket, bed, bike, bookshelf, \\ bridge, canvas, car, ceiling, \\ chair, corner, cylinder, dumbbells, \\ edge, fence, fitness equipment, glass, \\ ground, gun, handrail, heap, \\ hole, horse, jumping pit, \\ motorcycle, mud, obstacle, pad, \\ person, plank, platform, pole, \\ rock, rope, ropeswing, roundabout, \\ scooter, seasaw, shovel, skateboard, \\ slope, snow, springboard, stair, \\ stilts, stool, straw, swing, \\ toy, trampoline, treadmill, tree, \\ tricycle, vaulting \\ box, wall, water\end{tabular} \\ \hline
\end{tabular}
\vspace{2mm}
\caption{Classes for failure event, agent and risky region in EF dataset.}
\label{table.class}
\end{table}

\begin{figure*}[t!]
\begin{center}
\includegraphics[width=1.0\textwidth]{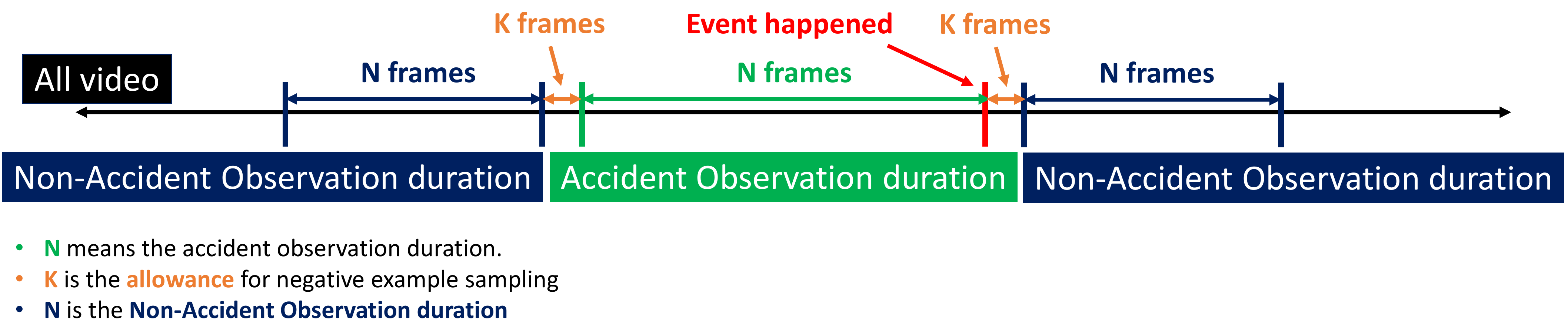}
\end{center}
\caption{Annotation protocol. The green color means accident video segment (Accident Observation duration) and the blue color means non-accident video segment (Non-Accident Observation duration). For consistent, we make accident video segment and the non-accident video segment in the same length (N frames). To make sure that non-accident video segment has enough margin to the accident video segment, we make N frames gap between them.}
\label{fig.anno}
\vspace{2mm}
\end{figure*}

The annotation protocol for EF dataset can be found in Fig. \ref{fig.anno}. We first annotate positive examples (Accident Observation duration) with $N$ frames in videos. Further, to make sure that negative examples (Non-Accident Observation duration) have enough margin to the positive examples, we sample negative examples $K$ frames before or after the positive examples. The number of frames of negative examples is set as $N$, too. If those sampled negative examples have the apparent risk phenomenon, the annotator would reject it.

\begin{figure}[t!]
\begin{center}
\includegraphics[width=0.48\textwidth]{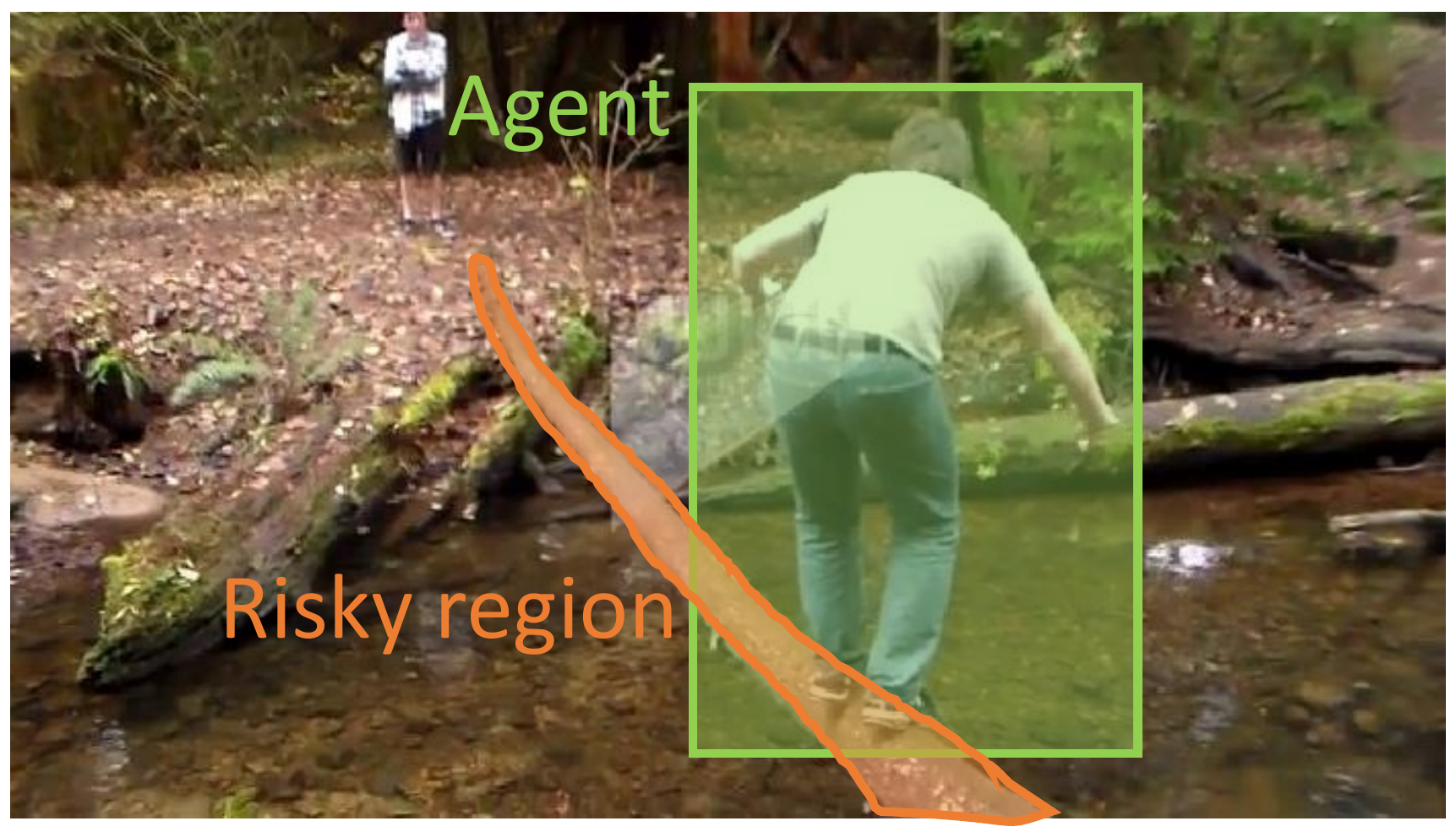}
\end{center}
\caption{Annotation for the agent and risky region. The green box shows the agent and the orange box shows the risky region.}
\label{fig.iSeg}
\vspace{-4mm}
\end{figure}

For each frame, we manually use bounding boxes and segmentation masks to annotate the agent (green color in Fig. \ref{fig.iSeg}) and risky region (orange color in Fig. \ref{fig.iSeg}). The whole annotation process is done with the iSeg tool \cite{schoningpixel}.
Even though with the help of the annotation tool, annotating is still time-consuming. Thus, testing data is carefully annotated at every frame, but training data is only annotated at every 15 frames.
We transform the segmentation maks for each risky region into a bounding box for evaluation purposes.

\section{Risk Map Visualization}
\label{E_risk_map}

The Agent-Region Interaction module of our framework outputs a risk score $s$ for each candidate risk region proposal. Since the risk score $s$ is produced by a sigmoid function, it corresponds to the risk probability for each proposal. Higher riskiness means that the proposal is riskier with respect to the agent. As a result, it raises an interesting question which is "Can we know the distribution of risk of the environment with respect to the agent?". We tackle it by utilizing riskiness of every proposal to visualize the risk map for each frame.

For the risk map shown in the qualitative results in the main paper and Fig. \ref{fig.vis_pos}, Fig. \ref{fig.vis_neg}, Fig. \ref{fig.vis_fail}, and Fig. \ref{fig.vis_SA}, we use the "imagesc" tool in Matlab \cite{MATLAB:2010} to draw risk maps. In detail, we generate risk maps by averaging the risk probability for each pixel over all covering candidate risky region proposals.
For visualization purposes, we standardize the color scale and range used to display the heatmaps. This enables easier comparison of heatmaps produced in different videos.
To do this, we first compute the highest risk probability in the testing set for each dataset. Then, we set it as the upper bound when we use the "imagesc" tool to generate risk maps. This setting helps us visualizing the results with bright color for higher risk probability and dark color for lower risk probability.

\section{Additional Qualitative Results}
\label{F_qualitative}

\paragraph{EF dataset}

\begin{figure*}[t!]
\begin{center}
\includegraphics[width=0.92\textwidth]{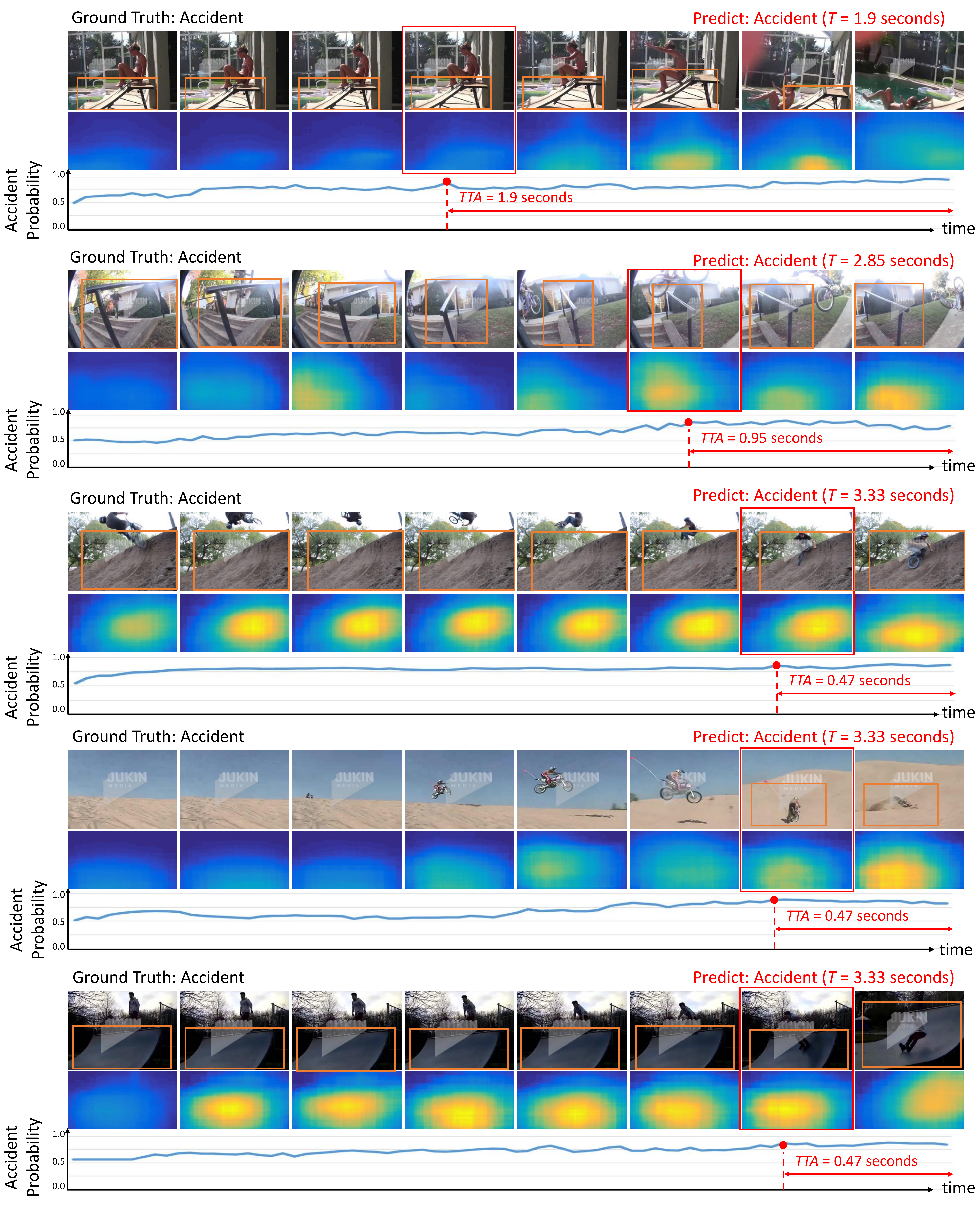}
\end{center}
\vspace{-2mm}
\caption{Qualitative Results for positive examples. We set 0.9 as the threshold for triggering accident anticipation. The accident anticipation probability (bottom row), the heat map (yellow for high risk and blue for low risk) for the risky regions (middle row) and ground truth risky region (orange box in top row) are shown in each example. For risk map, we average risk confidences of covering boxes for each pixel and draw the map by using Matlab [19] "imagesc" tool.}
\label{fig.vis_pos}
\end{figure*}

\begin{figure*}[t!]
\begin{center}
\includegraphics[width=0.92\textwidth]{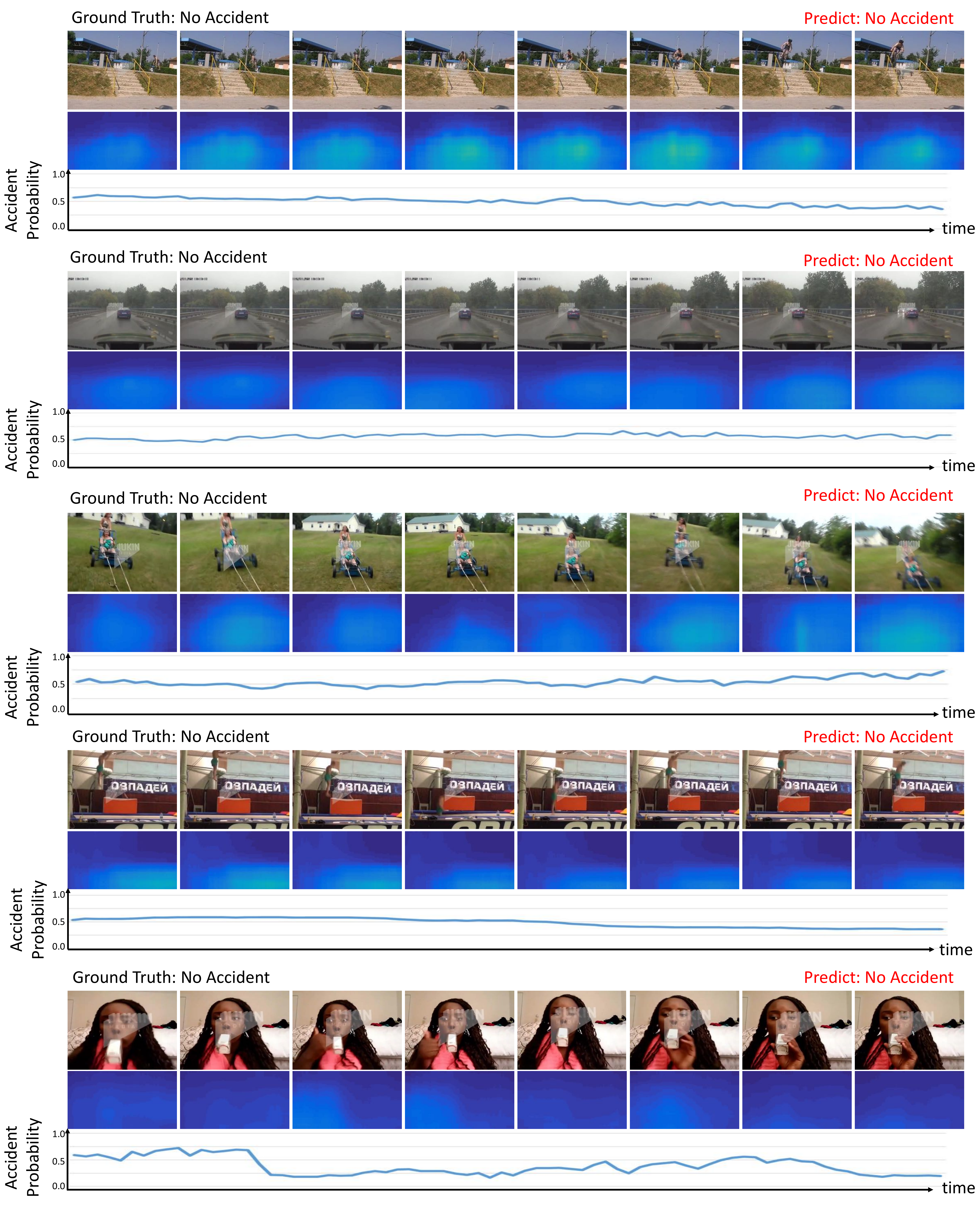}
\end{center}
\vspace{-2mm}
\caption{Qualitative Results for negative examples. For negative examples, the settings and the arrangement of the figure are the same as the positive examples in Fig. 4. We can clearly see that the colors of the risk map here are dim so that the model can easily recognize there is no accident event involved in the videos.}
\label{fig.vis_neg}
\end{figure*}

\begin{figure*}[t!]
\begin{center}
\includegraphics[width=0.92\textwidth]{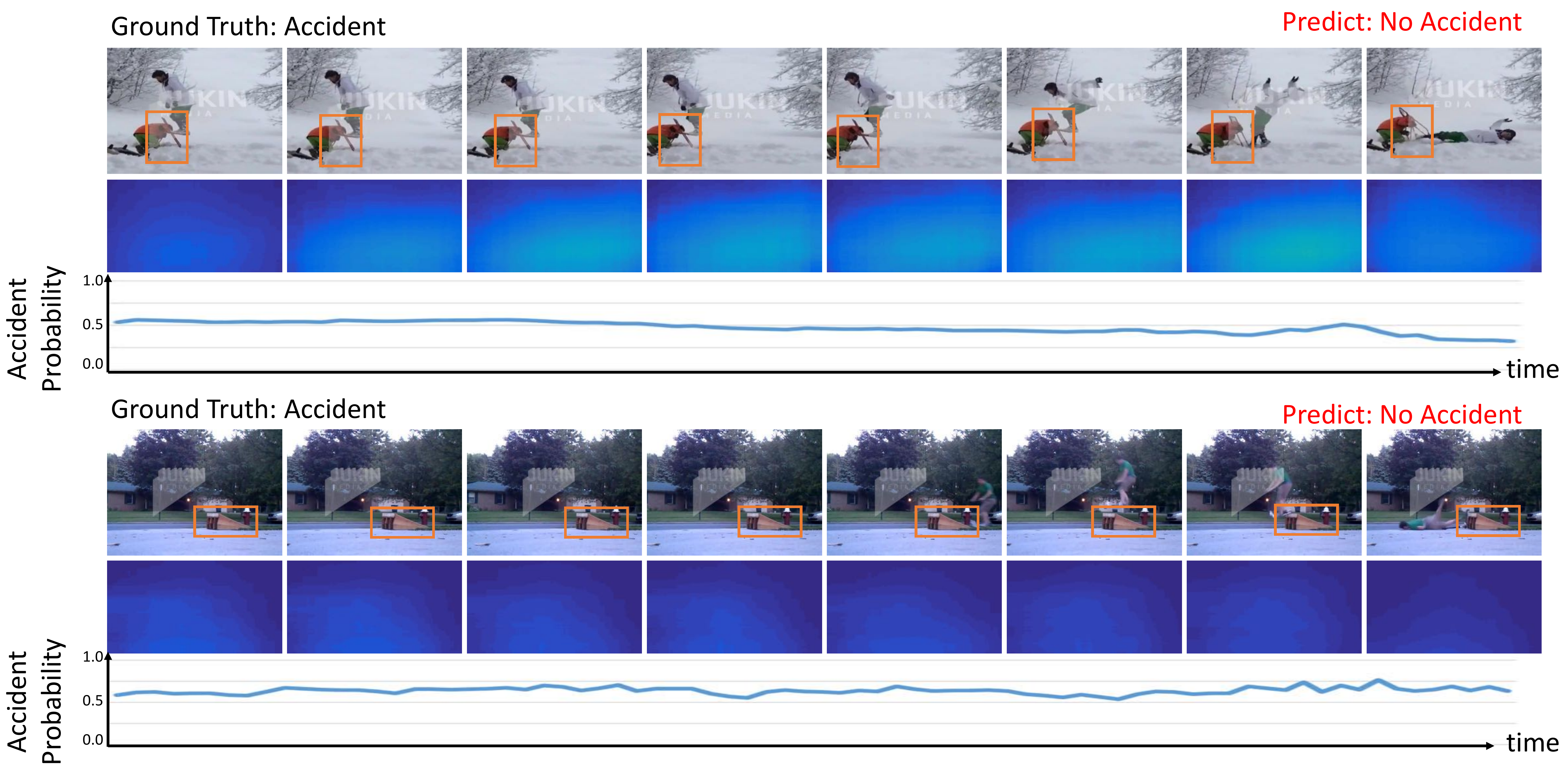}
\end{center}
\vspace{-2mm}
\caption{Qualitative Results for failure examples. The reason for failure for the first one is that the model does not recognize the risky region because the risky region (a chair) is similar to the sticks in the background. The reason for failure for the second one is that the color of the agent is similar to the background (the house in brown color). Therefore, the model confuses to recognize where is the risky region and results in the wrong prediction.}
\label{fig.vis_fail}
\end{figure*}

We show more qualitative results in Fig. \ref{fig.vis_pos}, Fig. \ref{fig.vis_neg} and Fig. \ref{fig.vis_fail}, where are results for positive examples, negative examples and failure examples. In the positive examples and some failure examples, we use orange bounding boxes to show the annotated risky region. We also show the $TTA$ measurement for all the positive examples and some failure examples. We describe the detail of the examples in the caption.

\paragraph{SA dataset}

\begin{figure*}[t!]
\begin{center}
\includegraphics[width=0.92\textwidth]{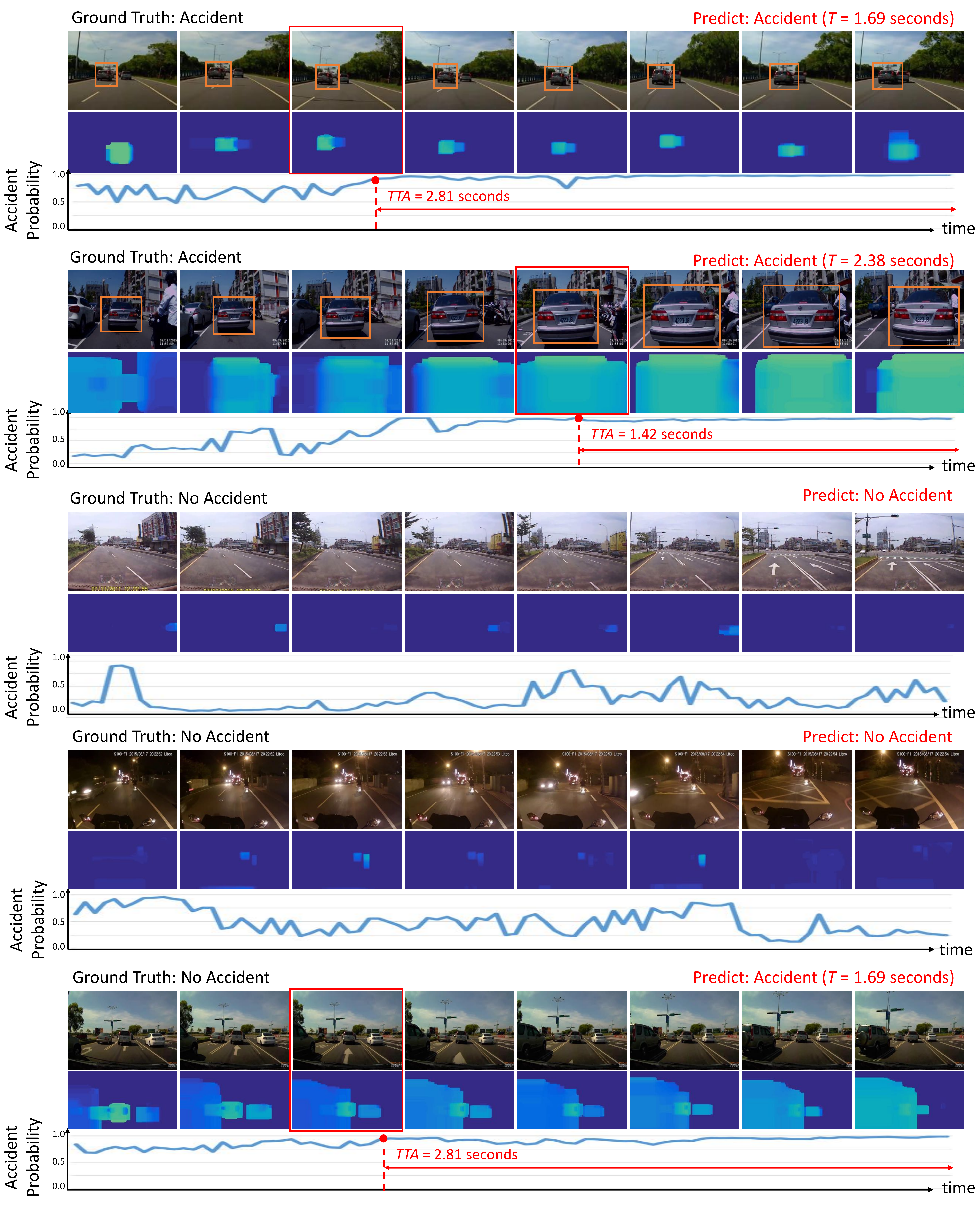}
\end{center}
\vspace{-2mm}
\caption{Qualitative Results for SA dataset. The first and second one are positive examples. The third and the fourth one are negative examples. The last one is the failure case. The reason for the failure case is that the three cars in the front are too close so that the model may attempt to recognize one of them is risky region.}
\label{fig.vis_SA}
\end{figure*}

We show qualitative results for SA dataset in Fig. \ref{fig.vis_SA}. The first and second one are positive examples. The third and fourth one are negative examples. The last one is a failure case. We describe the detail of the examples in the caption.

\end{document}